\pdfoutput=1

\documentclass[11pt]{article}

\usepackage{emnlp2021}

\usepackage{times}
\usepackage{latexsym}

\usepackage[T1]{fontenc}

\usepackage[utf8]{inputenc}

\usepackage{microtype}

\usepackage{microtype}
\usepackage{graphicx}
\usepackage{subfigure}
\usepackage{booktabs} 

\usepackage{multirow}
\usepackage{longtable}
\usepackage{amsmath}
\usepackage{tabularx}
\usepackage{dsfont}

\RequirePackage{algorithm}
\RequirePackage{algorithmic}

\DeclareMathOperator*{\argmax}{argmax}

\title{A Self-Paced Mixed Distillation Method for Non-Autoregressive Generation}

\author{Weizhen Qi\textsuperscript{1} \thanks{ \hspace{2mm}Work is done during internship at Microsoft Research Asia.}, 
Yeyun Gong\textsuperscript{2} \thanks{ \hspace{2mm}Corresponding Author.}  , 
Yelong Shen\textsuperscript{3}, 
Jian Jiao\textsuperscript{3}, 
Yu Yan\textsuperscript{3}, \\
\textbf{ Houqiang Li\textsuperscript{1} ,
Ruofei Zhang\textsuperscript{3} , 
Weizhu Chen\textsuperscript{3}, 
Nan Duan\textsuperscript{2}}   \\
  \textsuperscript{1}University of Science and Technology of China, \textsuperscript{2}Microsoft Research Asia,  \textsuperscript{3}Microsoft \\
  \texttt{\textsuperscript{1}weizhen@mail.ustc.edu.com, lihq@ustc.edu.com}, \\ 
   \texttt{\textsuperscript{2}\{yegong,nanduan\}@microsoft.com,  }\\  
  \texttt{\textsuperscript{3}\{yeshe,jian.jiao,yyua,bzhang,wzchen\}@microsoft.com } 
  }

\begin{document}
\maketitle
\begin{abstract}

Non-Autoregressive generation is a sequence generation paradigm, which removes the dependency between target tokens. It could efficiently reduce the text generation latency with parallel decoding in place of token-by-token sequential decoding. However, due to the known multi-modality problem, Non-Autoregressive (NAR) models significantly under-perform Auto-regressive (AR) models on various language generation tasks. Among the NAR models, BANG is the first large-scale pre-training model on English un-labeled raw text corpus. It considers different generation paradigms as its pre-training tasks including Auto-regressive (AR), Non-Autoregressive (NAR), and semi-Non-Autoregressive (semi-NAR) information flow with multi-stream strategy. It achieves state-of-the-art performance without any distillation techniques. However, AR distillation has been shown to be a very effective solution for improving NAR performance. In this paper, we propose a novel self-paced mixed distillation method to further improve the generation quality of BANG. Firstly, we propose the mixed distillation strategy based on the AR stream knowledge. Secondly, we encourage the model to focus on the samples with the same modality by self-paced learning. The proposed self-paced mixed distillation algorithm improves the generation quality and has no influence on the inference latency. We carry out extensive experiments on summarization and question generation tasks to validate the effectiveness. To further illustrate the commercial value of our approach, we conduct experiments on three generation tasks in real-world advertisements applications. Experimental results on commercial data show the effectiveness of the proposed model. Compared with BANG, it achieves significant BLEU score improvement. On the other hand, compared with auto-regressive generation method, it achieves more than 7x speedup. We will make our code publicly available.

\end{abstract}

\section{Introduction}
Non-AutoRegressive (NAR) models have been studied recently for efficient sequence generation~\cite{qi2021bang, gu2017non}. Different from classical Autoregressive (AR) approaches which sequentially decode output tokens~\cite{lewis2019bart, song2019mass, brown2020language,zou2021controllable,he2021petgen}, NAR approaches generate the sequence of tokens in parallel i.e. BANG ~\cite{qi2021bang}, NAT~\cite{gu2017non} etc, to largely reduce the inference latency, which have been successfully applied in query generation, text summarization tasks ~\cite{rajpurkar2016squad, narayan2018don, rush2015neural}. 

Despite reducing the inference time dramatically, typical NAR models still significantly under-perform AR models  ~\cite{qi2021bang}. 
Previous works analyze the issue of performance degradation by NAR and attribute it to the multi-modality problem ~\cite{kim2016sequence}. The multi-modality problem in NAR is described as generating target tokens from different possible answers and composing a chaotic confusing target sequence. It is not observed in AR models because they would pick only one possible answer with step-by-step generation, with all previous generated tokens as known information. To alleviate the multi-modality problem, sequence distillation~\cite{kim2016sequence, gu2017non} is widely used to replace the original training targets with the generated sequences by a well-trained AR model. Sequence distillation is analyzed to prove its ability to improve NAR performance by reducing the modality~\cite{zhou2019understanding} and reducing the dependency between target sequence tokens~\cite{ren2020study}. Besides sequence distillation, various techniques are proposed to improve the NAR generation including copy mechanism for translation~\cite{gu2017non}, curriculum learning ~\cite{guo2020fine}, glancing sampling~\cite{qian2020glancing}, pre-training~\cite{qi2021bang} etc. 

In this paper, we propose a novel self-paced mixed distillation method. Firstly, we propose to instruct the NAR model to select one modality to converge and focus on the samples with the same modality. At the beginning, NAR model will study all samples equally, then gradually select the easy samples with self-paced learning. We propose to use perplexity (PPL) to measure the modality-matching quality, and give rewards to the samples that agree with the converged modality. Secondly, we propose to generate soft labels from the BANG AR stream for teaching NAR stream. With the soft labels including rare words knowledge from original golden data rather than directly adding original data into training, it is less possible to hurt the NAR performance with increased modality problem. On the contrary, if we say the learned AR model regulates the data distribution to generalize a simplified fitting function, instead of the hard outputs from AR models which are approximately sampled from beam search, directly predicted words distribution better describe the AR learned generation function. The AR teacher model is trained on original golden data but teaches the student NAR model soft labels with distilled data as contexts. Experimental results show that the proposed mixed distillation and self-paced learning significantly improve NAR performance.

The contributions of this paper can be summarized as:
\begin{enumerate}
  \item We propose a self-paced mixed distillation method to teach BANG NAR generation with soft labels knowledge from its AR knowledge with self-paced learning.
  \item We carry out extensive experiments on summarization, question generation with obvious improvements. It is easy to deploy with significant performance improvements and no influence on inference latency.
  \item We applied the proposed method to commercial tasks. It achieves significantly performance improvement compared with BANG NAR. Compared with AR models, the proposed method meets the online requirement and also achieves comparable performance.  
\end{enumerate}

\section{Preliminary}
\subsection{Non-AutoRegressive Generation}
Consider the sequence to sequence generation scenario, we denote the input and output sequence as $(\mathbf{x}, \mathbf{y})$. For a typical neural sequence generation model, i.e., ~\cite{lewis2019bart, song2019mass, qi2020prophetnet}, it encodes the input sequence $\mathbf{x}$ into dense representation $\mathbf{h}$ in Eqn.~\ref{eq.encoder}, and decodes a sequence of tokens as output $\mathbf{y}:\{y_t\}^{T}_{t=1}$.
\begin{equation}
    \begin{aligned}
        \mathbf{h} & = \text{Encoder}(\mathbf{x}) 
    \end{aligned}
    \label{eq.encoder}
\end{equation}
In the classical Auto-Regressive generation (AR) paradigm ~\cite{brown2020language}, each token $y_i$ in the output sequence $\mathbf{y}$ is predicted with the dependency of $\mathbf{h}$ and previous tokens $\mathbf{y}_{<t}$, as in Eqn. ~\ref{eq.AG}.
\begin{equation}
    \begin{aligned}
        y_t = \text{Decoder}_{\text{AR}}(\mathbf{y}_{<t},  \mathbf{h}) 
    \end{aligned}
    \label{eq.AG}
\end{equation}
Non-AutoRegressive generation (NAR) models ~\cite{gu2017non, qi2021bang} predict each token $y_t$ of $\mathbf{y}$ simultaneously, given $\mathbf{h}$ and position $t$ in Eqn.~\ref{eq.NAG}. 
\begin{equation}
    \begin{aligned}
        y_t = \text{Decoder}_{\text{NAR}}(t, \mathbf{h}) 
    \end{aligned}
    \label{eq.NAG}
\end{equation}
NAR could greatly reduce the inference complexity compared with AR by discarding the dependency between sequence of output tokens. However, it degrades the performance of AR by introducing the multi-modality issue ~\cite{zhou2019understanding}. 

\subsection{BANG: Bridging Autoregressive and Non-autoregressive Generation}

\begin{figure}[h]
    \centering
    \includegraphics[width = 3.0in]{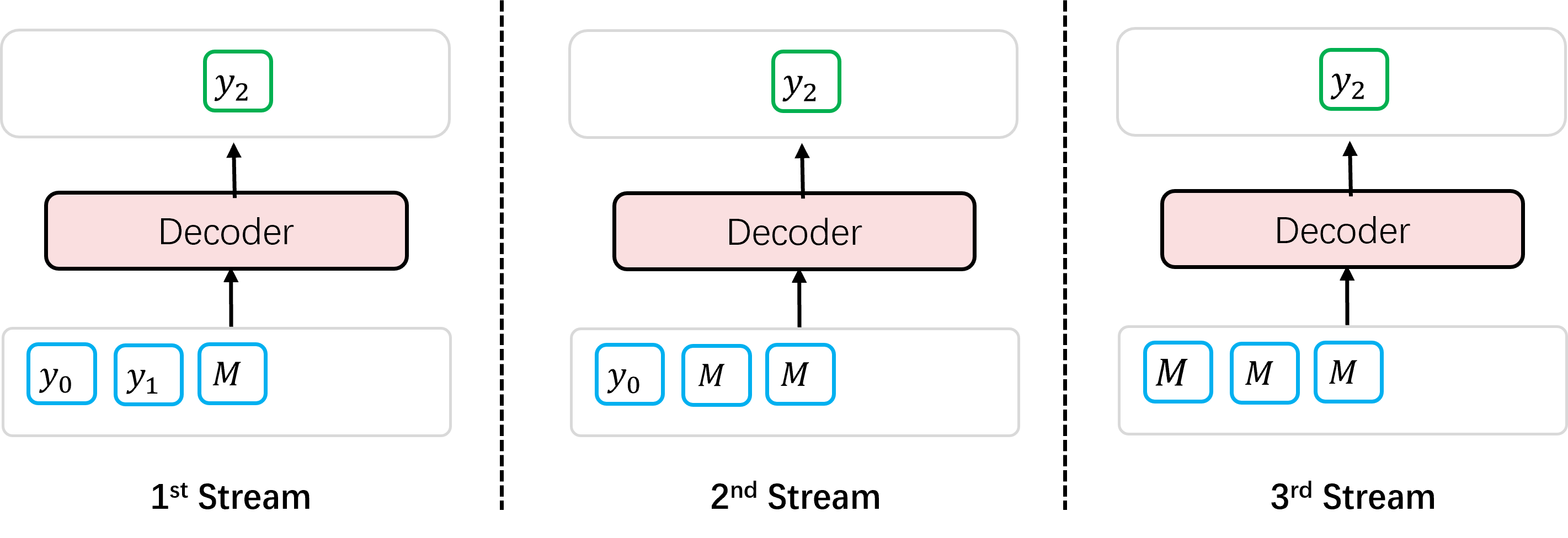}
	\caption{Three-Stream BANG model. In this example, M is short for [MASK] token. In $i^{th}$ predicting stream, $i-1$ previous tokens are masked out for AR/NAR generation. }
	\label{fig.bang.v1.ps}
\end{figure}

\textbf{BANG} ~\cite{qi2021bang} is a large-scale pre-trained language model with transformer based encoder-decoder architecture. It adopts $n$-stream self-attention mechanism for integrating AR, NAR and Semi-NAR generation paradigms into a unified model. In Figure ~\ref{fig.bang.v1.ps}, it illustrates a three-stream BANG model. The $1^{\text{st}}$ stream in BANG can be utilized for AR generation, and $2^{\text{nd}}$ and $3^{\text{rd}}$ streams are used for NAR/Semi-NAR generation.

 The conditional probabilities of generating target sequence $\mathbf{y}$ given $\mathbf{x}$ are shown in Eqn.~\ref{eq.BANG.v1}, where $\mathbf{p}^\mathbf{1}(\mathbf{y} | \mathbf{x})$ and $\mathbf{p}^\mathbf{n}(\mathbf{y} | \mathbf{x})$ indicate the conditional probabilities computed by the $1^\text{st}$ and $n^\text{th}$ streams in BANG, respectively. 
\begin{equation}
    \begin{aligned}
       \mathbf{p}^\mathbf{1}(\mathbf{y} | \mathbf{x}) &= \prod_{t=1}^{T}  p^\mathbf{1}(y_t |  \mathbf{y}_{<t}, \mathbf{x}) \\
        \mathbf{p}^\mathbf{2}(\mathbf{y} | \mathbf{x}) &= \prod_{t=1}^{T}  p^\mathbf{2}(y_t |  \mathbf{y}_{<t-1}, \mathbf{x}) \\
       \cdot&\cdot\cdot \\
        \mathbf{p}^\mathbf{n}(\mathbf{y} | \mathbf{x}) &= \prod_{t=1}^{T}  p^\mathbf{n}(y_t |  \mathbf{y}_{<t-n+1}, \mathbf{x}) \\
    \end{aligned}
    \label{eq.BANG.v1}
\end{equation} 
The pre-training objective for BANG minimizes the negative log-likelihood of target sequences for all the $n$ prediction streams, as in Eqn.~\ref{eq.BANG.v1.obj}. 
\begin{equation}
    \begin{aligned}
    \mathcal{L}_{\text{BANG}} (\mathbf{x}, \mathbf{y}) = -\sum_{s=1}^{n} \text{log} \mathbf{p}^s(\mathbf{y} | \mathbf{x})
    \end{aligned}
    \label{eq.BANG.v1.obj}
\end{equation}
To compute the $n$ prediction streams efficiently, BANG adopts the Cross-stream Visible N-stream self-attention mechanism to  
obtain all the n-stream predictions with one forward pass.  Therefore, in an extreme case when $n \geq T$, BANG could decode all the output tokens in parallel with the NAR paradigm. 
\begin{equation}
    \begin{aligned}
        \mathbf{p}^\mathbf{NAR}(\mathbf{y} | \mathbf{x}) &= \prod_{t=1}^{T}  p^\mathbf{t}(y_t | \mathbf{x}) \\
    \end{aligned}
    \label{eq.BANG.v1.NAR}
\end{equation} 
In the work, we leverage BANG model architecture as the test-bed to study AR and NAR mechanisms for language generation. For the sake of simplicity, we denote the AR generation model in BANG as $\mathbf{p}^{\text{AR}}(\mathbf{y} | \mathbf{x})$ which is also named $\mathbf{p}^{1}(\mathbf{y} | \mathbf{x})$ in Eqn.~\ref{eq.BANG.v1}. 

\section{Method}
The vanilla BANG model optimizes the $n$-stream predictions independently during training, which would cause severe multi-modality issue for NAR generation~\cite{zhou2019understanding, kim2016sequence}. 
In this section, we first introduce the self-paced learning and mixed distillation, respectively. Then, we introduce mixed distillation used in BANG pre-training.

\subsection{Mixed Sequence Distillation}\label{sec.method.mixeddistillation}
\label{subsec.md}
Distillation approaches adopt the ``teacher-student'' learning paradigm, where the AR model in BANG serves as the ``teacher'' model and the NAR model is viewed as ``student'' models. 

Both teacher and student models make prediction over sequence of tokens,  the general distillation function for sequence generation models $\mathbf{p}^{\text{AR}}$ and $\mathbf{p}^{\text{NAR}}$ is given by Eqn.~\ref{eq.sd.v1}: 
\begin{equation}
\begin{aligned}
\mathcal{L}_{\text{Distill}}\left( \mathbf{p}^\text{AR}, \mathbf{p}^\text{NAR}, \mathbf{x} \right) &= 
D_{\text{KL}} \left( \mathbf{p}^\text{AR}(\cdot | \mathbf{x}) \parallel \mathbf{p}^\text{NAR}(\cdot | \mathbf{x}) \right)
\end{aligned}
\label{eq.sd.v1} 
\end{equation}
where $D_{\text{KL}}(\cdot)$ is Kullback-Leibler divergence, and $\mathbf{p}^\text{AR}(\cdot | \mathbf{x})$ and $\mathbf{p}^\text{NAR}(\cdot | \mathbf{x})$ define the probability distribution over all possible output sequences by teacher and student model respectively.

Since it is intractable to compute the Eqn.~\ref{eq.sd.v1} directly,  we study three alternative ways to approximate the general distillation loss function, to be elaborated as follows. 

\textbf{Sequence distillation}: Sequence distillation approximates the probability distribution over sequences by teacher model with the one-hot distribution, which is:
\begin{equation}
\begin{aligned}
    \forall_{\mathbf{y}\in \mathcal{Y}} \:\:\: \mathbf{p}^\text{AR}(\mathbf{{y}} | \mathbf{x}) \approx  
\begin{cases}
    1, & \text{if } \mathbf{{y}} = \argmax_{\mathbf{\hat{y}}} \mathbf{p}^\text{AR}(\mathbf{\hat{y}} | \mathbf{x}) \\
    0,              & \text{otherwise}
\end{cases}
\end{aligned}
\label{eq.seq.distill}
\end{equation}
where $\mathcal{Y}$ denotes the set of all possible output sequences. 
In practice, we use beam search decoding algorithm to obtain sequence $\mathbf{y}^{\text{bs}}$ to approximate the sequence with the maximum probability by AR model:
\begin{equation}
\begin{aligned}
\mathbf{y}^{\text{bs}} \approx \argmax_{\mathbf{\hat{y}}} \mathbf{p}^\text{AR}(\mathbf{\hat{y}} | \mathbf{x})
\label{eq.beam}
\end{aligned}
\end{equation}

By integrating Eqn.~\ref{eq.seq.distill} and ~\ref{eq.beam} into Eqn ~\ref{eq.sd.v1}, the distillation loss could be approximated by:
\begin{equation}
\begin{aligned}
\mathcal{L}_{\text{Distill}} \approx \mathcal{L}_{\text{Seq-Distill}}\left( \mathbf{p}^{\text{AR}}, \mathbf{p}^{\text{NAR}}, \mathbf{x} \right) &= - \log \mathbf{p}^{\text{NAR}}(\mathbf{y}^{\text{bs}} | \mathbf{x})
\label{eq.seq.distill.v2}
\end{aligned}
\end{equation}

According to the formulation ~\ref{eq.seq.distill.v2}, the distillation training process can be simply explained as: the student model $\mathbf{p}^{\text{NAR}}$ is trained with the sequence-to-sequence dataset generated by the teacher model $\mathbf{p}^{\text{AR}}$.

Despite the simplicity of the sequence distillation approach, it omits the token-wised probability distribution of the teacher model. Thus, another token-wised teacher forcing distillation approach is introduced here.

\textbf{Teacher-Forcing Distillation}: We first factorize the joint sequence probability $\mathbf{p}^\text{NAR}$ and $\mathbf{p}^\text{AR}$ in Eqn.~\ref{eq.sd.v1}.
\begin{equation}
\begin{aligned}
\mathcal{L}&_{\text{Distill}}  \left(\mathbf{p}^\text{AR}, \mathbf{p}^\text{NAR}, \mathbf{x} \right) = &\\
& \sum_{y \in \mathcal{Y}} \sum_{t=1}^{|\mathbf{y}|} 
\mathbf{p}^{\text{AR}}(\mathbf{y}_{1:t-1}|\mathbf{x})  \\ &
D_{\text{KL}} \left(p^\text{AR}(\cdot |  \mathbf{{y}}_{<t}, \mathbf{x}) \parallel p^\text{NAR}(\cdot | t, \mathbf{x}) \right)
\end{aligned}
\label{eq.tf.v3} 
\end{equation}
where $\mathbf{p}^{\text{AR}}(\mathbf{y}_{1:t-1}|\mathbf{x})$  gives the sequence probability of $\mathbf{y}_{1:t-1}$ by the teacher AR model. In the teacher-forcing distillation approach, it approximates the distribution $\mathbf{p}^{\text{AR}}(\mathbf{y}_{1:t-1}|\mathbf{x})$ with the one-hot distribution given by the ground-truth sequence $\mathbf{y}^*$:

\begin{equation}
\begin{aligned}
    \forall_{\mathbf{y}\in \mathcal{Y}} \forall_{t\leq |\mathbf{y}| } \:\:\: \mathbf{p}^\text{AR}(\mathbf{{y}}_{1:t-1} | \mathbf{x}) \approx  \\
\begin{cases}
    1, & \text{if } \mathbf{y}_{1:t-1} =  \mathbf{y}^*_{1:t-1}\\
    0,              & \text{otherwise}
\end{cases}
\end{aligned}
\label{eq.seq.distill.tf}
\end{equation}

Therefore, by combining the Eqn.~\ref{eq.seq.distill.tf} with Eqn.~\ref{eq.tf.v3}, the teacher-forcing distillation loss could be given as follows:

\begin{equation}
\begin{aligned}
\mathcal{L}&_{\text{Distill}} \approx \mathcal{L}_{\text{TF-Distill}}
\left(\mathbf{p}^\text{AR}, \mathbf{p}^\text{NAR}, \mathbf{x} \right) =\\
&\sum_{t=1}^{|\mathbf{y}^*|} D_{\text{KL}} \left(p^\text{AR}(\cdot |  \mathbf{{y}}^*_{<t}, \mathbf{x}) \parallel p^\text{NAR}(\cdot | t, \mathbf{x}) \right)
\end{aligned}
\label{eq.tf.v4} 
\end{equation}





\textbf{Mixed Sequence Distillation}: To leverage the advantage of both sequence-wise and token-wise distillation approaches, mixed sequence distillation instead uses $\mathbf{y}^\text{bs}$ for $\mathbf{p}^{\text{AR}}(\mathbf{y}_{1:t-1}|\mathbf{x})$ approximation with similar manner as in Eqn.~\ref{eq.seq.distill.tf}.
\begin{equation}
\begin{aligned}
    \forall_{\mathbf{y}\in \mathcal{Y}} \forall_{t\leq |\mathbf{y}| } \:\:\: \mathbf{p}^\text{AR}(\mathbf{{y}}_{1:t-1} | \mathbf{x}) \approx  \\ 
\begin{cases}
    1, & \text{if } \mathbf{y}_{1:t-1} =  \mathbf{y}^{\text{bs}}_{1:t-1}\\
    0,              & \text{otherwise}
\end{cases}
\end{aligned}
\label{eq.mixseq.distill.tf}
\end{equation}
Thus, the objective function by mixed sequence distillation is given as:
\begin{equation}
\begin{aligned}
\mathcal{L}&_{\text{Distill}} \approx \mathcal{L}_{\text{Mixed-Distill}}  \left(\mathbf{p}^\text{AR}, \mathbf{p}^\text{NAR}, \mathbf{x} \right) =\\
&\sum_{t=1}^{|\mathbf{y^{\text{bs}}}|} D_{\text{KL}} \left(p^\text{AR}(\cdot |  \mathbf{{y}}^{\text{bs}}_{<t}, \mathbf{x}) \parallel p^\text{NAR}(\cdot | t, \mathbf{x}) \right)
\end{aligned}
\label{eq.md.v5} 
\end{equation}

In Eqn.~\ref{eq.seq.distill.v2}, ~\ref{eq.tf.v4} and~\ref{eq.md.v5}, it gives objective functions of sequence distillation, teacher-forcing distillation and mixed sequence distillation respectively. In the model training, we combine the distillation loss with original objective function in BANG, thus the overall training objective is defined by:
\begin{equation}
\begin{aligned}
\mathcal{L}_{\text{Overall}} (\mathbf{x}, \mathbf{y}) =\mathcal{L}_{\text{BANG}} (\mathbf{x}, \mathbf{y}) + \\
\gamma \mathcal{L}_{\text{Distill}}\left(\mathbf{p}^\text{AR}, \mathbf{p}^\text{NAR}, \mathbf{x}) \right) \\
\end{aligned}
\label{eq.bs.distill} 
\end{equation}


\subsection{Self-Paced Learning}
\label{subsec.sp}
Denote the training corpus for sequence distillation learning to be \{$\mathbf{x}^1, ..., \mathbf{x}^C$\}. Classical training algorithms sample instances from the corpus according to the static uniform distribution. Curriculum learning adopts dynamic data sampling strategy during training ~\cite{zhu-etal-2021-combining-curriculum-learning, guo2020fine, qian2020glancing}. For example, it imitates taking well-designed easy-to-hard training courses, where ``easy''  instances are more likely to be sampled at early training stage, and ``hard'' instances are with higher sampling probabilities at late training stage. 


In the section, we introduce a self-paced curriculum learning strategy for sequence distillation. Instead of human-crafted training courses, self-paced learning utilizes posterior probability of the student model to calculate the weight of each instance during training. Generally, it assigns an extra weight to each training instance: $\{(\lambda^1, \mathbf{x}^1), ..., (\lambda^C, \mathbf{x}^C)\}$;  
$\lambda^i$ is the sampling weight of the $i$-th instance; which could reflect the ``easy/hard'' degree of the training case. 

Let $loss_i$ denote the distillation loss of the $i$-th instance:
\begin{equation}
\begin{aligned}
loss_i = \mathcal{L}_{\text{Distill}} \left(\mathbf{p}^\text{AR}, \mathbf{p}^\text{NAR}, \mathbf{x}^i \right)
\end{aligned}
\label{eq.selfpaced1} 
\end{equation}
$loss_i$ measures the discrepancy between teacher and student models for the $i$-th sample, and let $\lambda^i= \exp(-{loss_i} )$.
Intuitively, large value of $\lambda^i$ indicates the instance is easy for distillation learning, thus it is assigned with a larger weight. In the practice of the self-paced learning, we adopt the batch-wise weight normalization to stabilize the training procedure. Thus, batch-wised self-paced distillation loss is computed by :
\begin{equation}
\begin{aligned}
\mathcal{L}_{\text{SP-Distill}}\left( \mathbf{p}^{\text{AR}}, \mathbf{p}^{\text{NAR}}, \{ (\lambda^i, \mathbf{x}^i \}_{i=1}^{B} \right) &=  \\
\sum_{i=1}^{B} \frac{\exp{\lambda^i}}{\sum_{o=1}^{B}{\exp{\lambda^o}}} \mathcal{L}_{\text{Distill}} \left(\mathbf{p}^\text{AR}, \mathbf{p}^\text{NAR}, \mathbf{x}^i \right) 
\end{aligned}
\label{eq.batch.selfpaced} 
\end{equation}

\subsection{Large Scale Pre-training}\label{sec.method.largescalepretrain}


In previous section ~\S~\ref{sec.method.mixeddistillation}, we introduced different distillation methods to teach the NAR training with AR knowledge. BANG has a list of predicting streams that can predict tokens in AR, semi-NAR or NAR information flow for pre-training. We propose to use $L_{TF-Distill}$ as a self-distillation method for further pre-training in larger corpus with nearly no extra cost. The same workflow is used for training self-distillation BANG as previous work, except that the training targets for NAR streams are replaced with the predicted distributions from AR stream. The algorithm is described in Alg~\ref{alg.bangv2}.

\begin{algorithm}[h]
   \caption{Large Scale Pre-training with Self-Distillation.}
   \label{alg:example}
\begin{algorithmic}
   \STATE {\bfseries Require:} Corpus $\mathcal{C}$; Distillation weight $\alpha$; Initialize the model with BANG.
   \FOR{article $\mathcal{A}$ in  get\_articles($\mathcal{C}$)  } 
   \STATE $noised\_article, spans = mask\_spans(\mathcal{A})$
   \STATE $x, y \leftarrow make\_batch(noised\_article, spans)$ \;
   \STATE $\hat{y} = BANG(x, \theta)$\;
   \STATE $\hat{y}^1, \hat{y}^2, ..., \hat{y}^n = split\_streams(\hat{y})$ \;
   \STATE $y_{soft} = \alpha y + (1-\alpha) \hat{y}^1.detach()$  \;
   \STATE $loss = mean(NLL(y, \hat{y}^1), $ \;
   \STATE $\qquad KL(y_{soft}, \hat{y}^2), ..., KL(y_{soft}, \hat{y}^n) )$ \;
   \STATE $\theta  \leftarrow loss.backward() $
   \ENDFOR
   \STATE return $\theta$ \;
\end{algorithmic}
\label{alg.bangv2}
\end{algorithm}

In Algorithm~\ref{alg.bangv2}, we can see the procedure to prepare training samples is the same as BANG. Given an article, a span of continues tokens is masked out to predict in the decoder, while the noised article is fed into the encoder as inputs. $\hat{y}$ is predicted from BANG multiple stream decoders. For $\hat{y}^i$ in i-th stream, tokens are predicted with $i-1$ previous tokens replaced with [MASK]. In another word, tokens in first stream $\hat{y}^1$ are predicted AR information flow. Each predicting stream will predict a distribution with different context to predict the same sequence. The distribution of AR stream will be used to calculate NLL loss with the golden hard targets. The predicted distribution of other predicting streams will be used to calculate KL divergence loss with the AR stream predictions.

\section{Experiments}\label{sec.experiments}

\subsection{Benchmarks}   
\subsubsection{Public Datasets}

We evaluate the proposed method on three publicly available benchmarks: SQuAD 1.1, XSum, and Gigaword for question generation and summarization tasks.

\textbf{SQuAD 1.1}~\cite{rajpurkar2016squad} is a question generation dataset, with 98K training samples. The data is formatted as $\langle$passage, answer, question$\rangle$. Each passage can be combined with various answers to raise different questions. We follow previous work~\cite{qi2020prophetnet, qi2021bang} to feed $\langle$answer [SEP] passage$\rangle$ into transformer encoder as the input, with an average length 149.4. The average output length is 11.5.  

\textbf{XSum}~\cite{narayan2018don} is a summarization dataset, with 204K training samples, 11K validation samples, and 11K test samples. Each sample includes an British Broadcasting Corporation (BBC) article and a professionally written single sentence summary. The average output length is 21.1.

\textbf{Gigaword}~\cite{rush2015neural} is a summarization dataset, containing 3.8M  training pairs, 189k validation pairs, and 1951 test pairs of $\langle$passage, summary$\rangle$ examples. They are extracted and cleaned from the Gigaword corpus~\cite{graff2003gigaword}. To be specific, it is a headline generation task, with the first sentence of the article as passage input, and the headline as summary. The average output length is 9.7.

\subsubsection{Real World Benchmarks}

We also deploy our proposed model on real world sponsored search engine applications. For a sponsored search engine, advertisers will provide their websites and their interested keywords, where keywords can also be auto-generated with a trained  landing page title-to-keyword generation model. When search engine users search a query, it has chances to trigger some keywords that advertisers have interest on, and the trigger procedure can be seen as a query-to-keywords generation task. We collect three commercial datasets for advertisements query-to-keyword generation and landing page title-to-keyword generation tasks.  The corpus was collected from En-US market. The corpus size of each dataset is shown in Table~\ref{corpus.ads}. The definition and collection details are as following:

\begin{table}[h]
\small \centering
\caption{The corpus size of QKG-EM, QKG-BM, and ATKG datasets.}
\begin{tabular}{l|ccc|c}\hline
Dataset & Train &  Valid  &   Test  & All \\ \hline
QKG-EM &  72,876   &   10,000 & 2,130  &    85,006   \\ 
QKG-BM &   6,474,865  &  10,000  & 492,278 &  6,977,143  \\ 
ATKG &  5,001,037   & 10,000   & 355,824 &  5,366,861     \\  \hline
\end{tabular}\label{corpus.ads}
\end{table}

\begin{table*}[h]
\small \centering
\caption{The performance of our methods and baseline methods for non-autoregressive summarization task on XSum benchmark. ``(+x.xx)'' means the absolute improvement based on BANG.}
\begin{tabular}{lccccc}\hline
MODEL  & PRE-TRAIN  & ROUGE-1 & ROUGE-2 & ROUGE-L&  OVERALL	 \\ \hline
NAT~\cite{gu2017non}    & No     & 24.04   & 3.88 & 20.32 &   16.08  \\ 
CMLM~\cite{ghazvininejad2019mask}    & No        & 23.82   & 3.60    & 20.15 &   15.86 \\ 
LevT~\cite{gu2019levenshtein}    & No        & 24.75  & 4.18    & 20.87 &   16.60 \\ 
BANG ~\cite{qi2021bang}   & Yes       & 32.59  & 8.98    & 27.41 &   22.99 \\ \hline
BANG + $\mathcal{L}_{\text{SP}}$ & Yes  & 33.01(+0.42)  & 9.27(+0.29)   & 27.76(+0.35)  &  23.35(+0.36) \\
BANG  + $\mathcal{L}_{\text{TF-Distill}}$   & Yes     & 34.72 (+2.13)  & 10.18 (+1.20)  & 29.36 (+1.95)&   24.75 (+1.76)\\ 
BANG + $\mathcal{L}_{\text{SP-TF-Distill}}$    & Yes    & 35.02(+2.43)  & 10.37(+1.39)  & 29.52(+2.11) &   24.97(+1.98)  \\ 
BANG  + $\mathcal{L}_{\text{BS-Hard-Distill}}$   & Yes     & 35.22 (+2.63)  & 11.82(+2.84) & 29.36(+1.95) &   25.47(+2.48) \\ 
BANG  + $\mathcal{L}_{\text{BS-Distill}}$   & Yes     & 36.13 (+3.54)  & 11.73 (+2.75)  & 30.02 (+2.61)&   25.96 (+2.97)\\ 
BANG  + $\mathcal{L}_{\text{SP-BS-Distill}}$  & Yes  & 36.26 (+3.67) & 12.04(+3.06)  & 30.19 (+2.78)&  \textbf{26.16(+3.17)} \\  
\hline
\end{tabular}\label{table.main.XSum.v1}
\end{table*}

\begin{table*}
\small \centering
\caption{Non-autoregressive generation performance on Gigaword summarization. SD is short for sequence distillation, with the AR distilled training set. Soft means with training with AR predicted soft lables. self-paced means reverse self-paced learning with training samples re-weighting.}
\begin{tabular}{lccccc}\hline
Model   & PRE-TRAIN   & ROUGE-1 & ROUGE-2 & ROUGE-L&   OVERALL 	 \\ \hline
BANG  ~\cite{qi2021bang}     & Yes       & 32.61  & 13.39 & 30.76  &   25.59 \\ \hline
BANG + $\mathcal{L}_{\text{SP}}$ & Yes  &  33.09(+0.48)   & 14.12(+0.73)  & 31.30(+0.54)  & 26.17(+0.58)  \\
BANG + $\mathcal{L}_{\text{TF-Distill}}$    & Yes         & 33.30 (+0.69) &  14.01 (+0.62)  &  31.38(+0.62) &   26.23 (+0.64) \\ 
BANG + $\mathcal{L}_{\text{SP-TF-Distill}}$    & Yes    &  33.75(+1.14)  & 14.50(+1.11)  & 31.80(+1.04)  &   26.68(+1.09)  \\ 
BANG + $\mathcal{L}_{\text{BS-Hard-Distill}}$     & Yes    & 36.13(+3.52)  & 16.95(+3.56) & 33.75(+2.99) &  28.94 (+3.35) \\
BANG + $\mathcal{L}_{\text{BS-Distill}}$  & Yes     &  36.32 (+3.71)   & 17.28(+3.89) & 34.04 (+3.28) &   29.21 (+3.62) \\ 
BANG  +  $\mathcal{L}_{\text{SP-BS-Distill}}$  & Yes    &  36.62(+4.01)  &  17.74(+4.35) & 34.29(+3.53) &  \textbf{29.55(+3.96)} \\ 
\hline
\end{tabular}\label{table.main.gigaword}
\end{table*}

\textbf{QKG-EM}: Query to close variant keywords generation for exact match. In this task, given a user query, the model generates a list of keywords that have exactly the same intent as the source query. 
Such a situation usually occurs when advertisers have a clear targeted audience, judging from the search queries. To construct QKG-EM, we collect the user query and keywords from clicked ads. Then, three crowdsourcing annotators are asked to give a binary label for each query and keyword pair. We determine the data label when more than two annotators reach a consensus. The average target sequence length in the training set and test set is 3.21 and 2.52 respectively. After tokenization into word pieces, the numbers are 4.07 and 3.42.
        
\textbf{QKG-BM}: Query to keywords generation for broad match. In this task, given a user query, the model generates a list of keywords that is semantic relevant to the query. This happens when advertisers want to reach to a broader slice of users that may be interested in their product. Similar to construct QKG-EM, we collect a set of query and keyword pairs from clicked data. And because QKG-BM is harder to judge, we ask five crowdsourcing annotators to label each pair of QKG-BM. When more than three people reach a consensus, we determine the final label. The average target sequence length in the training set and test set is 2.70 and 2.94 respectively. After tokenization into word pieces, the numbers are 3.68 and 3.91.
        
\textbf{ATKG}: Ad title to keywords generation. In this task, given an ad landing page title, the model generates a list of keywords that are relevant to the ad title. For many electronic business platforms, there are lots of products without ready-made keywords of ad. This task tends to automatically generate keywords. To construct ATKG, we collect query and landing page title pairs through clicked data, and regard the query as the keywords of the landing page title. Then, three crowdsourcing annotators are asked to label each pair, and we also determine the final label by consensus.  The average target sequence length in the training set and test set is 3.71 and 4.04 respectively. After tokenization into word pieces, the numbers are 4.77 and 5.28.

For these tasks, the AR models latency can not meet the requirements while optimized NAR generation model can be  online used to meet the real time usage.

\subsection{Baselines}

We cite the NAR baseline model results from ~\citet{qi2021bang}. The referred baseline models include: \textbf{NAT}~\cite{gu2017non}, \textbf{CMLM}~\cite{ghazvininejad2019mask}, \textbf{LevT}~\cite{gu2019levenshtein}, and \textbf{BANG}~\cite{qi2021bang}. NAT is the first non-autoregressive translation model based on Transformer, it removes the unidirectional information flow constraint and introduces sequence distillation, target length prediction, decoder inputs copy techniques. CMLM predicts arbitrary subset of masked words in a target sequence with the masked language model objective. LevT adopts insertion and deletion as basic operations to edit the draft. BANG is our most related NAR model and has been thoroughly introduced. We follow ~\citet{qi2021bang} to cite the first round outputs of CMLM and LevT, NAR finetuning results of BANG as their NAR results. We carry out improvements on the base of BANG. The BANG variants with our proposed techniques are notated as:

\textbf{BANG+{\text{TF-Distill}}}: It uses the teacher-forcing distillation method for enhancing the model training, as described in Section ~\S~\ref{subsec.md}. In short words, soft labels with original training data serving as previous tokens.

\textbf{BANG+{\text{BS-Distill}}}: It uses the beam-search distillation method in the model training, as described in Section ~\S~\ref{subsec.md}. In short words, soft labels with beam search output training data serving as previous tokens.

\textbf{BANG+{\text{BS-Hard-Distill}}}: It also uses the beam-search distillation method, but instead of using the predicting score of the autoregressive teacher model for distillation, it uses one-hot vector for distillation, this kind of distillation method have been widely used in non-autoregressive models~\cite{gu2017non}.

\textbf{BANG+{\text{SP-BS-Distill}}}: It combines the self-paced learning for teacher-forcing distillation, as described in Section ~\S~\ref{subsec.sp}.

\subsection{Main Results}\label{experiments}

\begin{table*}[h]
\small \centering
\caption{Non-autoregressive generation performance on SQuAD 1.1 question generation. SD is short for sequence distillation, with the AR distilled training set. Soft means with training with AR predicted soft lables. self-paced means reverse self-paced learning with training samples re-weighting.}
\begin{tabular}{lccccc}\hline
MODEL  & PRE-TRAIN  & ROUGE-L & BLEU-4 & METEOR &  OVERALL \\   \hline
NAT~\cite{gu2017non}  & No           & 31.51  & 2.46 & 8.86  &  14.29 \\ 
CMLM~\cite{ghazvininejad2019mask}     & No        & 32.44   & 2.33 & 8.84  &   14.54  \\ 
LevT~\cite{gu2019levenshtein}     & No        & 31.38  & 2.27 & 9.14  &   14.26 \\ 
BANG  ~\cite{qi2021bang}      & Yes     & 44.07 & 12.75 & 18.99 &  25.27 \\ \hline
BANG + $\mathcal{L}_{\text{SP}}$  & Yes    & 44.54 (+0.47)  & 13.61(+0.86)     & 19.46 (+0.47)   &  25.87(+0.60)  \\
BANG + $\mathcal{L}_{\text{TF-Distill}}$    & Yes     & 46.14(+2.07)  & 13.54(+0.79) & 20.06(+1.07)  &   26.58(+1.31) \\ 
BANG + $\mathcal{L}_{\text{SP-TF-Distill}}$    & Yes     & 46.49(+2.42)  & 14.14(+1.39) & 20.34(+1.35) &   26.99(+1.72) \\ 
BANG +  $\mathcal{L}_{\text{BS-Hard-Distill}}$   & Yes     & 46.14   (+2.07) & 15.19 (+2.44)  & 21.03 (+2.04)  &  27.45  (+2.18) \\ 
BANG + $\mathcal{L}_{\text{BS-Distill}}$    & Yes    & 47.26 (+3.19)  & 15.30  (+2.55) &  21.05  (+2.06) &   27.87 (+2.60)  \\ 
BANG  +  $\mathcal{L}_{\text{SP-BS-Distill}}$  & Yes    & 47.41 (+3.34)  & 15.64  (+2.89) & 21.22  (+2.23)   &   \textbf{28.09 (+2.82)}  \\ 
\hline
\end{tabular}\label{table.main.squad}
\end{table*}

We report the performance of our methods and baselines for non-autoregressive summarization task on XSum and Gigaword benchmarks 
in Table~\ref{table.main.XSum.v1} and~\ref{table.main.gigaword}.
 From the performance of ``BANG'' and ``BANG  + $\mathcal{L}_{\text{TF-Distill}}$'', we see that teacher forcing distillation achieves 1.76 and 0.64 points absolute performance improvement on overall score for XSum and gigaword. It illustrates strong autoregressive teacher model can help the non-autoregressive learning by soft labels knowledge without the beam search inference procedure. Comparing the performance of ``BANG'' and ``BANG+$\mathcal{L}_{\text{SP}}$'' we see the emphasis of easy samples will lead to a better converged model. Comparing the performance of ``BANG  + $\mathcal{L}_{\text{BS-Distill}}$'' with ``BANG  + $\mathcal{L}_{\text{TF-Distill}}$'' and ``BANG  + $\mathcal{L}_{\text{BS-Hard-Distill}}$'', we find that the proposed mixed distillation method achieves better performance than other distillation method. From the performance in Table~\ref{table.main.XSum.v1} and~\ref{table.main.gigaword}, we see that ``BANG  + $\mathcal{L}_{\text{SP-BS-Distill}}$'' achieves new state-of-the-art performance on both XSum and Gigaword benchmarks, and compared with BANG, it achieves 3.17 and 3.92 points absolute improvement, respectively. The results demonstrate the proposed self-paced mixed distillation method for non-autoregressive generation is effective.

In Table~\ref{table.main.squad}, we show the comparison of our methods and baselines on SQuAD 1.1 for question generation task. We reach conclusions consistent with summarization. ``BANG  + $\mathcal{L}_{\text{SP-BS-Distill}}$'' achieves new state-of-the-art performance and improve the the overall score 2.82 points.

\subsection{Ablation Study}

\subsubsection{Distillation with Soft versus Hard Target} In the section~\ref{subsec.md}, it presents the distillation learning with soft target by calculating the KL-divergence between the teacher and student models' predictions in Eqn.~\ref{eq.bs.distill}. 

\begin{table}[h]
\small \centering
\caption{The performance on SQuAD 1.1 of different $\gamma$ for $\mathcal{L}_{\text{TF-Distill}}$. OVL is short for OVERALL score.}
\begin{tabular}{lcccc}\hline
$\gamma =$     & ROUGE-L & BLEU-4 & METEOR & OVL \\ \hline
0.00     &  43.71  & 12.30  & 19.00 & 25.00  \\ 
0.25       & 43.86  &  12.33  & 19.18  & 25.12 \\ 
0.50       &  44.43  &  13.00  & 19.52  & 25.65 \\ 
0.75       &  45.26 &  13.52  & 20.07  & 26.28 \\ 
1.00  & 46.14  & 13.54 & 20.06  & 26.58  \\ \hline
\end{tabular}\label{table.qg.soft.weight}
\end{table}

We set a combination of hard and soft targets and show the results in Table~\ref{table.qg.soft.weight}. We reproduce the BANG NAR results and set all of the hyper-parameters the same(including the random seed), to equally compare the combination of hard and soft labels' weight. A consistent improvement can be seen when increasing the soft weight. It can be seen that soft labels are more suitable than hard labels for NAR learning.   

\subsubsection{Self-paced learning strategy}

In ~\S~\ref{subsec.sp}, we propose to focus on modality-consistency easy samples. Here we present the results if we focus on the hard samples:

\begin{table}[h]
\small \centering
\caption{BANG NAR results with different self-paced learning $\lambda_i$ for  $\mathcal{L}_{\text{SP}}$.  Here if $\lambda$ is set to None, then the model is same as BANG NAR. OVL is short for OVERALL score. }
\begin{tabular}{lcccc}\hline
$\lambda_i$=     & ROUGE-L & BLEU-4 & METEOR & OVL \\ \hline
loss   &   42.25  &  9.70     & 17.09     &  23.01   \\ 
log loss  &  42.88   &  10.45     &  17.75    &   23.69  \\ 
None     & 44.07   & 12.75  & 18.99 & 25.27 \\ 
1/PPL  & 44.54   & 13.61     & 19.46    &  25.87  \\ \hline
\end{tabular}\label{table.qg.sp.lamda1}
\end{table}

\begin{table}[h]
\small \centering
\caption{BANG NAR results with different self-paced learning $\lambda_i$ for $\mathcal{L}_{\text{SP-BS-Distill}}$. Here if $\lambda$ is set to None, then the model is same as $\mathcal{L}_{\text{BS-Distill}}$. OVL is short for OVERALL score. }
\begin{tabular}{lcccc}\hline
$\lambda_i$=     & ROUGE-L & BLEU-4 & METEOR & OVL \\ \hline
loss   & 46.94   &  15.03     & 20.85 &   27.61   \\ 
log loss  &   46.51   &  14.21   &  20.40    &  27.04  \\ 
None    & 47.26 & 15.30  &  21.05 &  27.87 \\ 
1/PPL& 47.41  & 15.64  & 21.22   &  28.09  \\ \hline
\end{tabular}\label{table.qg.sp.lamda2}
\end{table}

Comparison of how to calculating $\lambda_i$ is shown in Table~\ref{table.qg.sp.lamda1} and ~\ref{table.qg.sp.lamda2}. Here, $\lambda_i = PPL=exp(loss)$, $\lambda_i=loss$ and $\lambda_i=log(loss)$ is to focus on hard examples. $\lambda_i=1/PPL=1/exp(loss)$ is our proposed self-paced learning strategy. It can be observed  the hard examples focus sp strategies hurt the performance for both   $\mathcal{L}_{\text{TF-Distill}}$ in Table~\ref{table.qg.sp.lamda1} and $\mathcal{L}_{\text{SP-BS-Distill}}$ in  Table~\ref{table.qg.sp.lamda2}. It shows that the NAR models do not have the capacity to learn from hard multi-modality training samples, but the modality consistent easy data will help NAR models learn a fluent generation pattern.

\subsubsection{Non-AutoRegressive versus AutoRegressive generation}

The self-paced soft distillation has no influence on the inference latency, thus we cited the AR and NAR latency from ~\citet{qi2021bang} for readers that are not familiar with NAR performance. We list the Transformer AR performance and latency to be compared with BANG NAR model in Table~\ref{tab.transformervsbang.squad} and Table~\ref{tab.transformervsbang.xsum} for SQuAD 1.1 question generation and XSum summarization.

\begin{table}[h]
\small \centering
\caption{Latency (ms/sample) on SQuAD 1.1 question generation.  In this table, R-L, B-4, MTR are short for ROUGE-L, BLEU-4, and  METEOR respectively. }
\begin{tabular}{lcccc}\hline
MODEL   & R-L & B-4 & MTR  & LATENCY\\   \hline
Transformer &  29.43   & 4.61 &  9.86  & 159.49 \\ 
BANG     & 44.07 & 12.75 & 18.99  & 15.69 \\ 
+  $\mathcal{L}_{\text{SP-BS-Distill}}$   & 47.41 & 15.64  & 21.22     & 15.69  \\ 
\hline
\end{tabular}\label{tab.transformervsbang.squad}
\end{table}

\begin{table}[h]
\small \centering
\caption{Latency (ms/sample) on XSum summarization.  In this table,R-1,R-2,R-L are short for ROUGE-1, ROUGE-2, and ROUGE-L respectively. }
\begin{tabular}{lcccc}\hline
MODEL  & R-1 & R-2 & R-L  & LATENCY\\   \hline
Transformer &  30.66   & 10.80 &  24.48  & 262.47 \\ 
BANG    & 32.59 & 8.98 & 27.41 & 15.97 \\ 
+  $\mathcal{L}_{\text{SP-BS-Distill}}$   & 36.26  & 12.04  & 30.19    & 15.97  \\ 
\hline
\end{tabular}\label{tab.transformervsbang.xsum}
\end{table}

\subsubsection{Multi-stage Finetuning}
In previous sections, the NAR student model is initialized with the pre-trained model. Here we discuss initializing the NAR model with different starting points.

\begin{table}[h]
\small \centering
\caption{SQuAD 1.1 question generation results. In this table, R-L, B-4, MTR are short for ROUGE-L, BLEU-4, and METEOR respectively. }
\begin{tabular}{lccccc}\hline
Stage-1  & Stage-2       & R-L & B-4 & MTR  	 \\ \hline
 -  & NAR    & 44.07   & 13.61    & 19.46  \\
AR & NAR     & 44.77  & 13.00   & 19.62   \\ 
$\mathcal{L}_{\text{SP-BS-Distill}}$ & NAR   &  43.12  & 12.30   & 19.10   \\ \hline
 - &  $\mathcal{L}_{\text{SP-BS-Distill}}$   &  47.41 & 15.64 & 21.22  \\ 
NAR    & $\mathcal{L}_{\text{SP-BS-Distill}}$     & 46.71  &  15.16  &  20.95  \\ 
AR & $\mathcal{L}_{\text{SP-BS-Distill}}$   & 47.71  &  15.90  & 21.52   \\ 
$\mathcal{L}_{\text{BS-Hard-Distill}}$ & $\mathcal{L}_{\text{SP-BS-Distill}}$   &  47.25  & 15.58 & 21.12    \\ \hline
 -  & $\mathcal{L}_{\text{BS-Hard-Distill}}$  &  46.14 & 15.19 & 21.03   \\ 
NAR & $\mathcal{L}_{\text{BS-Hard-Distill}}$   & 45.96  &  14.90  & 20.79   \\ 
- & $\mathcal{L}_{\text{BS-Distill}}$   & 47.26 & 15.30   &  21.05   \\ \hline
\end{tabular}\label{tab.2stage}
\end{table}

In Table~\ref{tab.2stage}, we load different models before finetuning, as a two-stage training workflow. The two-stage finetuning experimental results help to claim these points:

1)  No need to specially train the samples equally before focusing on the easy samples with self-paced learning. Comparing the results of $\mathcal{L}_{\text{BS-Hard-Distill}}$ + $\mathcal{L}_{\text{SP-BS-Distill}}$, we find it's on par with directly $\mathcal{L}_{\text{SP-BS-Distill}}$ finetuning. It is because that although the modality consistency score is calculated with the PPL (or loss), when starting the training, the training samples' losses are very close and can be seen as equally learning, then gradually emphasize the easy samples.

2) Comparing the results of $\mathcal{L}_{\text{SP-BS-Distill}}$ with NAR+$\mathcal{L}_{\text{SP-BS-Distill}}$, and NAR with $\mathcal{L}_{\text{SP-BS-Distill}}$ + NAR, we see performance damage on both of the extra stage 1 pre-finetuning. It shows that the $\mathcal{L}_{\text{SP-BS-Distill}}$ reinforces the local optimization, while the converged NAR model on original data does not agree with the self-paced local optimal. The $\mathcal{L}_{\text{SP-BS-Distill}}$ will result in a better performance modality, which will not help the original training corpus.

3) Simply adding original training data will hurt sequence distillation performance, while adding original knowledge as soft distributions does not, when observing the performance of $\mathcal{L}_{\text{BS-Hard-Distill}}$, NAR + $\mathcal{L}_{\text{BS-Hard-Distill}}$ and $\mathcal{L}_{\text{BS-Distill}}$ . To benefit from original data, specific algorithms should be used~\cite{ding2021rejuvenating, ding2020understanding}, otherwise the performance may be damaged with the increased modality as our experimental results. Soft labels learning could be a simple yet effective choice to keep more information from raw data. 

4) It's interesting to find that by loading the parameters from AR teacher model, performance can be further improved for both NAR finetuning or $\mathcal{L}_{\text{SP-BS-Distill}}$ finetuning. It is probably because BANG structure supports different generation pattern naturally.

\subsubsection{Self-distillation to teacher NAR generation with shared parameters AR teacher}

In previous sections, the AR teacher models parameters are frozen after the AR finetuning procedure to act as a stable teacher. Next we want to validate that will soft labels distillation help NAR performance as a self-distillation strategy, then we can validate the effectiveness before employing it on large-scale pre-training.   Considering that all predicting streams of BANG share the model parameters during pre-training, here we carry out experiments to finetune a same model for both AR and NAR generation, with and without the knowledge from AR stream to NAR stream. 

We finetune a BANG model with 50\% batch of data in AR information flow and 50\% batch of data in NAR information flow on sequence distilled SQuAD 1.1 question generation benchmark, which we note as $\mathcal{L}_{\text{BS-Hard-Distill}}$.  We train another model with the same setting except that the NAR targets are AR predicted distributions and note as ${\text{BS-Soft-Self-Distill}}$.  The results are shown in Table~\ref{tab.qg.share}.

\begin{table}[h]
\small \centering
\caption{SQuAD 1.1 question generation. Infer is short for inference type. R-L, B-4, and MTR are short for ROUGE-L, BLEU-4, and METEOR, respectively.}
\begin{tabular}{lcccc}
\hline
Model   &  Infer   & R-L & B-4 & MTR  \\ \hline
$\mathcal{L}_{\text{BS-Hard-Distill}}$ & NAR & 45.98 & 14.87  & 20.65  \\
$\mathcal{L}_{\text{BS-Soft-Self-Distill}}$ & NAR & 46.41   & 15.25  & 20.91  \\ \hline
$\mathcal{L}_{\text{BS-Hard-Distill}}$ &   AR    & 46.77 & 18.18  & 22.09  \\ 
$\mathcal{L}_{\text{BS-Soft-Self-Distill}}$ & AR &  46.68    &  17.95  &  21.98  \\ \hline
\end{tabular}\label{tab.qg.share}
\end{table}

Comparing results in Table~\ref{tab.qg.share} and Tabel~\ref{table.main.squad} we can see that with the same model that able to generate outputs in both AR and NAR information flow(Table~\ref{tab.qg.share}), the outputs are slightly worse than directly NAR finetuing (Table~\ref{table.main.squad}). It is reasonable because the same model parameters are shared for different generation pattern. Comparing the NAR performance in Table~\ref{tab.qg.share} we can see the improvements by teaching knowledge from its AR stream. It motivates us to improve the NAR performance of BANG pre-training to use the AR stream predicted distributions for teaching other streams as introduced in section ~\S~\ref{sec.method.largescalepretrain}.

\subsection{Results for Real-World Advertisements Applications}\label{sec.real.world}

We show the results of BANG AR teacher model, BANG NAR baseline model and our improvements with $\mathcal{L}_{\text{SP-BS-Distill}}$ finetuning for three real world advertisements datasets in Table~\ref{table.results.EM}, Table~\ref{table.results.SM}, and Table~\ref{table.results.ATQ}. For AR teacher model, the beam size is set as 5 and length penalty as 1.2 for all the test set evaluation. Inference batch size is set to 1 to evaluate the latency to simulate online deployment. Notice that the final deployed BANG NAR generation model will be further optimized to accelerate, while for fair comparison, here we keeps the same code base as previous released BANG model. 

\begin{table}[h]
\small \centering
\caption{Performance and latency (ms/sample) on Query to Keywords Generation dataset QKG-EM. In this table, B- is short for BLEU-.}
\begin{tabular}{lcccc}\hline
Model       & B-1 & B-2 & B-4 & LATENCY \\ \hline
BANG AR &  61.27  &  48.90  & 31.02 & 120.48 \\  
BANG NAR &  67.07   &  55.76  & 28.35 & 16.69 \\  \hline
+$\mathcal{L}_{\text{SP-BS-Distill}}$ &  66.10   &  56.11  & 29.61 &  16.60 \\ \hline
\end{tabular}\label{table.results.EM}
\end{table}

\begin{table}[h]
\small \centering
\caption{Performance and latency (ms/sample) on Query to Keywords Generation dataset QKG-BM. In this table, B- is short for BLEU-.}
\begin{tabular}{lcccc}\hline
Model       & B-1 & B-2 & B-4 & LATENCY \\ \hline
BANG AR  &  38.04  &  27.12  & 6.15 & 115.74 \\ 
BANG NAR  &  31.53   &  17.15  & 2.59  & 17.16 \\  
+$\mathcal{L}_{\text{SP-BS-Distill}}$ & 37.25  & 26.58 &  6.14 & 16.60 \\ \hline
\end{tabular}\label{table.results.SM}
\end{table}

\begin{table}[h]
\small \centering
\caption{Performance and latency (ms/sample) on Ad landing page Title to Keywords Generation dataset ATKG. In this table, B- is short for BLEU-.}
\begin{tabular}{lcccc}\hline
Model       & B-1 & B-2 & B-4 & LATENCY \\ \hline
BANG AR  &  40.06   & 27.54   & 11.65 & 144.09 \\  
BANG NAR  &  28.19   &  21.61  & 8.17 & 16.73 \\  
+$\mathcal{L}_{\text{SP-BS-Distill}}$ & 39.38 &  26.91  & 11.41 & 16.96 \\ \hline
\end{tabular}\label{table.results.ATQ}
\end{table}

Obviously we can see the NAR generation will significantly reduce the inference latency, which can be deployed on real-world keywords extension usage. The difference between BANG NAR and $\mathcal{L}_{\text{SP-BS-Distill}}$ models can be ignored and resulted by the machine performance fluctuation because $\mathcal{L}_{\text{SP-BS-Distill}}$ has no effect on the inference procedure. For QKG-BM and ATKG, $\mathcal{L}_{\text{SP-BS-Distill}}$ reduces the performance gap between NAR model and AR teacher model significantly while keeps the same latency. It is exciting for sponsored search engine keywords extension tasks. Another interesting  observation is that for query to keywords extension QKG-EM, BANG NAR generation has better performance than AR generation for BLEU-1 and BLEU-2, while worse performance for BLEU-4. It shows that when the training data is not very adequate, meantime the output is short keywords, NAR generation is possible to outperform AR generation regarding single word and two adjacent words performance as BLEU-1 and BLEU-2, while still worse performance regards relatively longer fluent expresstions as BLEU-4. With $\mathcal{L}_{\text{SP-BS-Distill}}$, the BLEU-4 score is improved while the BLEU-1 and BLEU-2 is hurt, which means that our proposed method will make the NAR student model more consistent with the AR teacher model rather than simply improving evaluation metrics. Generally speaking, with our proposed learning method, BANG NAR model has satisfying performance close to AR generation but much lower latency.

\subsection{Pre-training Results}

We perform further pre-training on 160GB unlabeled English corpus, including news, books, stories and web text. It is similar to the corpus of well-known AR pre-training works such as ProphetNet~\cite{qi2020prophetnet} and BART~\cite{lewis2019bart}. The learning rate is set to 4e-4, 366k steps, batch size 2048,  distillation weight $\alpha$ 0.5 on 16 32GB memory NVIDIA Tesla V100 GPUs. We show the reusults for XSum summarization and SQuAD 1.1 question generation in Table~\ref{table.160g.XSum} and Table~\ref{table.160g.squad}.

\begin{table}[h]
\small \centering
\caption{Non-autoregressive generation performance on XSum summarization.  BANG$_{160g}$ means our pretrained model to initialize the model before finetuning. Teacher models are the same for fair comparison.}
\begin{tabular}{lccccc}\hline
Pretrain  & Finetune       & R-1 & R-2 & R-L 	 \\ \hline
BANG   & NAR    & 32.59   & 8.98    & 27.41  \\ 
BANG$_{160g}$ & NAR     & 33.55   & 9.69    & 28.30  \\
BANG$_{160g}$ & $\mathcal{L}_{\text{SP-BS-Distill}}$   & 36.65 & 12.70  & 30.61  \\ \hline
\end{tabular}\label{table.160g.XSum}
\end{table}

\begin{table}[h]
\small \centering
\caption{Non-autoregressive generation performance on SQuAD 1.1 question generation. BANG$_{160g}$ means our pretrained model to initialize the model before finetuning. Teacher models are the same for fair comparison.}
\begin{tabular}{lcccc}\hline
Pretrain & Finetune        & R-L & B-4 &MTR 	 \\ \hline
BANG & NAR & 44.07   & 12.75  & 18.99 \\ 
BANG$_{160g}$ & NAR  & 44.59   & 12.97  & 19.55  \\ 
BANG$_{160g}$ & $\mathcal{L}_{\text{SP-BS-Distill}}$   & 47.83 & 16.20  & 21.59 \\ \hline
\end{tabular}\label{table.160g.squad}
\end{table}

We can see that with self-distillation further pre-training, performance is consistently improved among the two benchmarks and different NAR finetuning methods. To ensure the results comparable, the teacher model for 160 $\mathcal{L}_{\text{SP-BS-Distill}}$ finetuning keeps the same as BANG $\mathcal{L}_{\text{SP-BS-Distill}}$ baseline. We will also release the further pretrained model when our code is open sourced.

\section{Related Work}
AR generation has been widely developed in recent years, and pre-training techniques achieve significantly performance improvement in AR generation tasks~\cite{GPT3,BART,T5,qi2020prophetnet}. GPT3~\cite{GPT3} pre-train a large model and generate the next token from left-to-right. BART~\cite{BART}, T5~\cite{T5}, and ProphetNet~\cite{qi2020prophetnet} are based on encoder-decoder architecture. BART~\cite{BART} pre-train the model through reconstructing the original text from a noised input. ProphetNet~\cite{qi2020prophetnet} learn to recover a mask span of a input text with a n-gram prediction mechanism. T5~\cite{T5} investigates different pre-training techniques and pre-train a generation model with large scale corpus. Pre-training techniques are well-developed in AR generation tasks. 

Different from AR generation, few pre-training works focus on NAR generation. BANG~\cite{qi2021bang} is the first large scale pre-training work for NAR generation. It combines AR, NAR, and semi-NAR in the pre-training. Except pre-training, sequence distillation is one powerful method to improve the performance in NAR generation. It has been widely studied~\cite{gu2017non,zhou2019understanding,ren2020study}.~\citet{zhou2019understanding} analyze sequence distillation from reducing the modality perspective. 
And~\citet{ren2020study} study it from reducing the dependency between target sequence tokens perspective. Besides sequence distillation, glancing sampling~\cite{qian2020glancing}, curriculum learning from AR model~\cite{guo2020fine}, and encoder copy for translation~\cite{gu2017non} are proposed to reduce the difficulty of NAR generation.

In this work, we propose a new self-paced mixed distillation method to reduce the difficulty of NAR generation and successfully applied it
to BANG.

\section{Conclusion}

In this paper, we propose several techniques to improve the non-autoregressive generation performance based on BANG. Firstly, we propose to use mixed distillation to keep the knowledge from original corpus rather than completely ignoring them or simply adding them back. Secondly, self-paced learning is adopted to focus on the easy samples for modality-consistent. Then we extend the mixed distillation into self-distillation pre-training for BANG to utilize its autoregressive stream knowledge. Extensive experiments are carried out to support our claims. We see significant improvements on the public benchmarks including summarization tasks XSum and gigaword, question generation tasks SQuAD 1.1. We also deploy our model in real-world sponsored search engine applications.

\bibliography{anthology,custom}
\bibliographystyle{acl_natbib}




\end{document}